%% file: paper.tex
\documentclass[runningheads]{llncs}
\usepackage[T1]{fontenc}
\usepackage{graphicx}
\usepackage{booktabs}
\usepackage{multirow}
\usepackage{xspace}
\usepackage[misc]{ifsym}

\input{definitions.tex}

\usepackage{subcaption}
\usepackage{colortbl}

\newcommand{\shadeOP}[2]{%
  \ifdim #1pt<0.853pt
    \cellcolor[gray]{#1}#2%
  \else
    #2%
  \fi
}
\newcommand{\shadeBS}[2]{%
  \ifdim #1pt<0.864pt
    \cellcolor[gray]{#1}#2%
  \else
    #2%
  \fi
}
\newcommand{\bX}{\ensuremath{\boldsymbol{x}}\xspace}
\newcommand{\by}{\ensuremath{\boldsymbol{y}}\xspace}
\newcommand{\bT}{\ensuremath{\boldsymbol{\theta}}\xspace}
\newcommand{\reprModel}{\phi_{\boldsymbol{\theta}}}
\newcommand{\trainedModel}{f_{\boldsymbol{\theta}}}
\newcommand{\numberdatasets}{five\xspace}

\begin{document}

\title{Dual-Criterion Curriculum Learning: Application to Temporal Data}

\titlerunning{Model-driven Dual-Criterion Curriculum Learning for Temporal Data}

\author{%
Gaspard Abel\inst{1,2}\ \
Eloi Campagne\inst{1,3}\ \
Mohamed Benloughmari\inst{1}\\
Argyris Kalogeratos\inst{1}
}



\institute{
Université Paris Saclay, Université Paris Cité, ENS Paris Saclay, CNRS, SSA, INSERM, Centre Borelli, F-91190, Gif-sur- Yvette, France
\and
Centre d'Analyse et de Mathématique Sociales, EHESS, CNRS,
75006 Paris, France
\and
EDF R\&D, 91120 Palaiseau, France\\
\email{\{name.surname\}@ens-paris-saclay.fr}
}

\maketitle              
\begin{abstract}
Curriculum Learning (CL) is a meta-learning paradigm that trains a model by feeding the data instances incrementally
according to a schedule, which is based on difficulty progression.
Defining meaningful difficulty assessment measures is crucial and most usually the main bottleneck for effective learning, while also in many cases the employed heuristics are only application-specific.
In this work, we propose the Dual-Criterion Curriculum Learning (DCCL) framework that combines two views of assessing instance-wise difficulty: a loss-based criterion is complemented by a density-based criterion learned in the data representation space.
Essentially, DCCL calibrates training-based evidence (loss)
under the consideration that data sparseness amplifies the learning difficulty.
As a testbed, we choose the time-series forecasting task. We evaluate our framework on multivariate time-series benchmarks under standard One-Pass and Baby-Steps training schedules.
Empirical results show the interest of density-based and hybrid dual-criterion curricula over loss-only baselines and standard non-CL training in this setting.
\end{abstract}


\section{Introduction and Related Work}
\label{sec:Intro_Rel}

While classical training feeds a model with all data instances at once, Curriculum Learning (CL)~\cite{bengioCurriculumLearning2009} is a paradigm for sequential model training: instances are presented in a sequence that goes typically from easy to hard, mimicking the way humans and animals progressively acquire knowledge over a matter~\cite{leeWhyAnimalsNeed2024}. This technique has been explored theoretically~\cite{weinshallCurriculumLearningTransfer2018,sagliettiAnalyticalTheoryCurriculum2022} as well as empirically where it has been shown to improve convergence speed and generalization in various domains, most notably in Computer Vision~\cite{guiCurriculumLearningFacial2017,guoCurriculumNetWeaklySupervised2018} and Natural Language Processing~\cite{spitkovskyBabyStepsLeapfrog2010,taySimpleEffectiveCurriculum2019}, while in applications like Robotics this principle can be adapted by designing tasks that are increasingly far from the desired target \cite{florensaReverseCurriculumGeneration2017}.
An alternative to keeping model complexity fixed and complexifying the input data (or task) is to perform the opposite: the sequential complexification of the model architecture~\cite{karrasProgressiveGrowingGANs2018}. We refer the reader to surveys~\cite{sovianyCurriculumLearningSurvey2022,liuReviewEvaluationSystem2023} for an alternative overview of the different CL definitions.

According to the definition from \cite{wangSurveyCurriculumLearning2022}, with a fixed model complexity, two distinct design modules for CL are formulated: a \emph{difficulty measurer}, which assigns a difficulty score to each data instance, and a \emph{training scheduler}, which determines how those scores are translated into a sequence of training sets. Common choices for the difficulty measurer range from predefined heuristics (\eg sequence length in language processing~\cite{spitkovskyBabyStepsLeapfrog2010}, image intensity~\cite{guiCurriculumLearningFacial2017}) to automatic approaches such as Self-Paced Learning~\cite{kumarSelfPacedLearningLatent2010}, where difficulty is inferred from the model's own training loss.
For the training scheduler, first of all, the difficulty progression is typically discretized by splitting the dataset in \emph{buckets}. The two canonical strategies are the \emph{One-Pass} that exposes the model to each bucket exactly once in order of increasing difficulty, and the \emph{Baby-Steps}~\cite{spitkovskyBabyStepsLeapfrog2010} that augments the training set by including more buckets to avoid forgetting easy instances. Formalizing the difficulty measurer via transfer from a pre-trained model is explored in \cite{weinshallCurriculumLearningTransfer2018}, providing both a practical recipe and theoretical convergence guarantees.

In a rather limited line of work, density estimation has been used as a principled proxy for instance difficulty in several supervised curriculum learning works. The idea is that instances in denser regions are more promising to be easy for the learning model. Density is a data characteristic that can be measured in a \emph{problem-agnostic} way in the input data, or in a \emph{problem-aware} way by considering embeddings created by generic or the very model to be learned.
For instance, CurriculumNet \cite{guoCurriculumNetWeaklySupervised2018} obtains image representations from a pretrained InceptionV2 model and defines difficulty by the local density of those representations: noisy images fall into low-density regions and are therefore scheduled later in training. A similar intuition underlies for unsupervised domain adaptation in \cite{choiPseudoLabelingCurriculumUnsupervised2019}, where high-density regions in the representation space correspond to high-confidence pseudo-labels and are thus treated as easy instances. These methods operate in a supervised or pseudo-supervised setting and are applied to classification tasks. Even fewer works have investigated ways to exploit data density in a problem-agnostic and less supervised way, \eg by using clustering to capture the density in spectral data embeddings and dynamically adjusting the sampling weights of easy and hard instances for the intent detection task \cite{gongDensityBasedDynamicCurriculum2021}, or by other simpler clustering-based heuristics \cite{chaudhryDataDistributionBasedCurriculum2024}.

\inlinetitle{Curriculum Learning for Time-Series Analysis}{.}
In dynamical systems analysis some works have introduced curriculum learning to combine teacher forcing and free running \cite{teutschFlippedClassroomEffective2022}, which demonstrate gains on chaotic dynamical systems. Similarly, \cite{bucciCurriculumLearningDatadriven2023} applies curriculum learning to data-driven forecasting of dynamical systems, using the entropy of the output distribution as an automatic difficulty measure. In the financial domain, \cite{koeneckeCurriculumLearningDeep2019} defines instance difficulty via the residual error obtained from Seasonal Trend decomposition using Loess (STL)~\cite{cleveland1990stl}, and builds a curriculum for LSTM and dilated CNN forecasters, providing accuracy improvements over standard forecasting models. For multivariate settings, \cite{wuConnectingDotsMultivariate2020} proposes a GNN approach that captures inter-series dependencies, though without an explicit curriculum component.
Overall, despite the widespread CL applications in a series of fields, its adoption in temporal data analysis remains particularly limited. This motivates our application focus therein, yet without restricting our methodological approach that is intended to be general.

\inlinetitle{Contributions}{.}
This work is primarily motivated by the fact that defining difficulty assessment measures is crucial and is usually the main bottleneck for effective CL, as well as that in many cases the employed heuristics are only application-specific. Our contributions can be summarized as follows:
\begin{itemize}
\item  We propose the Dual-Criterion Curriculum Learning (DCCL), a modular framework that combines two views of assessing instance-wise learning difficulty: a loss-based criterion, complemented by a density-based criterion learned in a data representation space.
DCCL calibrates training-based evidence about difficulty, essentially captured by the loss, under the consideration that data sparseness amplifies the learning difficulty. To the best of our knowledge, this is the first work putting forward the idea that data density and loss can be regarded as complementary criteria to jointly determine difficulty progression in the context of CL.

\item Our dual-criterion formulation is generic and allows the definition of hybrid CL strategies. We propose ways to address challenging methodological issues, notably: how to estimate the data density in a suitable embedding space for the data in question (\eg such as Transformers or LSTMs), and how to fuse the two criteria. 
The wealth of representation learning methods for typical vector data, but also for more complex structured data such as temporal data or graphs, render the DCCL framework applicable in a wide range of data types and tasks.

\item Our application focus is on time-series analysis, which is relatively unexplored in the CL literature. We evaluate CL strategies on Transformers for multivariate time-series forecasting, providing insights on the relative benefits of curricula on an up-to-date architecture. Our framework is evaluated on benchmark datasets for time-series forecasting, and effectively outperforms standard CL strategies.
\end{itemize}

\section{Preliminaries on Curriculum Learning}
\label{sec:cl_general}

In this work, we see the CL pipeline according to the definition presented in \cite{wangSurveyCurriculumLearning2022}, namely composed of two main components: a \emph{difficulty measurer} that assigns a difficulty score to each instance, and a \emph{training scheduler} that uses those scores to determine sequences of training sets for the model. We also consider presenting the data to a model in the typical easy-to-hard difficulty order.

\inlinetitle{Difficulty measurer}{.}~%

A \emph{difficulty function},
  $\delta : \mathcal{X} \longrightarrow \mathbb{R}$,
assigns a scalar difficulty score $\delta(\bX_n)$ to each training instance
$\bX_n \in \mathcal{D}$.
A higher value of $\delta$ indicates a harder instance. Note that the difficulty of an instance depends on the complexity of the learning problem at hand and the amount and quality of the available data in $\mathcal{D}$. The choice of $\delta$ 
is the 
major design choice. The difficulty measurer may be \emph{predefined}, such as sequence length in NLP \cite{spitkovskyBabyStepsLeapfrog2010,taySimpleEffectiveCurriculum2019}, or noise intensity~\cite{guoCurriculumNetWeaklySupervised2018} and signal intensity~\cite{guiCurriculumLearningFacial2017} in image processing. Another common choice is to use an \emph{automatic} assignment of difficulty, such as the loss of a model (\eg Self-Paced Learning~\cite{kumarSelfPacedLearningLatent2010}), or Reinforcement Learning Teacher frameworks~\cite{weinshallCurriculumLearningTransfer2018}, where a teacher model assigns difficulty scores to instances for training a student model. In this work, we focus on a model-driven approach to difficulty measurement, based on density estimation in a learned representation space.

\inlinetitle{Training scheduler}{.}~%
Any difficulty criterion $\delta$ assigning a scalar score per data instance provides an empirical density function over that score (\ie a histogram of score counts). Then, CL can be formulated as an \emph{filtration} that maps different strata of the density to each curriculum bucket.
Given all scores $\{\delta(\bX_n)\}_{n=1}^{N}$, the dataset is partitioned into $K$ ordered \emph{curriculum buckets}
\[
  \mathcal{B}_1 \preceq \mathcal{B}_2 \preceq \cdots \preceq \mathcal{B}_K,\quad \text{where}\ \ \mathcal{D} =	{\textstyle \bigcup_{k=1}^{K}} \mathcal{B}_k,
\]
and $\preceq$ denotes the easy-to-hard ordering: $\mathcal{B}_1$ contains the easiest instances (lowest difficulty scores) and $\mathcal{B}_K$ the hardest.
The model $f_{\bT}$ is then trained in $K$ successive stages:
at stage $k$, $f_{\bT}$ is fine-tuned on a \emph{training set} $\mathcal{C}_k$ built from the buckets according to a chosen \emph{schedule}, starting from the parameters $\bT^{(k-1)}$ obtained at the previous stage. Formally,
\[
  \bT^{(k)} \;=\; \operatorname*{arg\,min}_{\bT}
  \;\mathcal{L}\!\left(\bT;\, \mathcal{C}_k\right),
  \quad k = 1, \ldots, K,
\]
with $\bT^{(0)}$ being a pre-trained initialization.
The \emph{schedule}, \ie the rule that maps stage index $k$ to training set $\mathcal{C}_k$, is a distinct design choice to the difficulty measurer.
Two standard choices are:
\begin{itemize}[leftmargin=1.2em,itemsep=0.4em,topsep=.4em]
  \item \emph{One-Pass.} At each stage $k$, the model is refined exclusively on the new bucket, $\mathcal{C}_k = \mathcal{B}_k$, which is seen
		exactly once, in order. This schedule is fast, but may suffer from forgetting of earlier seen, easier instances.

  \item \emph{Baby-Steps.} At each stage $k$, the model is refined on all the buckets seen so far: $\mathcal{C}_k \;=\; \bigcup_{j=1}^{k} \mathcal{B}_j$. 
    Hence, the model is re-exposed to all previously seen instances, which stabilizes training at the cost of a growing training set. 
\end{itemize}

It should be noted that most \emph{automatic} CL approaches, such as Self-Paced Learning~\cite{kumarSelfPacedLearningLatent2010}, do not explicitly predefine a training scheduler, since the curriculum is implicitly defined by the model's learning dynamics.
In this work, our focus is set on the \emph{difficulty measurer} component of the curriculum learning pipeline, whereas the \emph{training scheduler} is kept fixed to the standard One-Pass and Baby-Steps schedules. This allows us to isolate and demonstrate better the effects of a difficulty measure on the performance of the CL framework.

\section{A Framework for Curriculum Learning}
\label{sec:CL_TS}

In this section, we first present the proposed DCCL Curriculum Learning framework and detail its modular components, then we describe density-based and loss-based difficulty measures, and finally we introduce dual-criterion variants that combine both those measures.

\subsection{The DCCL Curriculum Learning Pipeline}
\begin{figure}[t]
\centering
\resizebox{\linewidth}{!}{%
\begin{tikzpicture}[
  box/.style={
    draw, rounded corners=4pt, thick,
    minimum width=2.7cm, minimum height=1.3cm,
    align=center, font=\small, text width=2.0cm, inner sep=5pt
  },
  biggerbox/.style={
    draw, rounded corners=4pt, thick,
    minimum width=2.1cm, minimum height=1.3cm,
    align=center, font=\small, text width=2.5cm, inner sep=5pt
  },
  bigggerbox/.style={
    draw, rounded corners=4pt, thick,
    minimum width=3cm, minimum height=1.3cm,
    align=center, font=\small, text width=2.5cm, inner sep=5pt
  },
  modbox/.style={
    box, dashed, draw=blue!65, fill=white
  },
  pipearrow/.style={->, >=Stealth, line width=1.3pt},
  sublbl/.style={font=\scriptsize, align=center},
  node distance = 1.0cm
]


\node[biggerbox, fill=blue!12] (encoder) {
  \textbf{Representation Model}\\[3pt]
  $\reprModel:\mathcal{X}\to\mathbb{R}^{d}$\\[1pt]
};

\node[modbox, right=1.1cm of encoder] (diff) {
  \textbf{Difficulty Measurer}\\[3pt]
  $\delta:\mathbb{R}^d\!\to\!\mathbb{R}$\\[1pt]
  {$\{\delta_n:=\delta(\reprModel(\bX_n))\}_{n=1}^{N}$}
};

\node[bigggerbox, fill=blue!10, right=1.1cm of diff] (adap_filter) {
  \textbf{Adaptive Filtering}\\[3pt]
  $\{\delta_n\}_{n=1}^{N} \to \{\mathcal{B}_1\preceq\cdots\preceq\mathcal{B}_K\}$\\[1pt]
  {\footnotesize cluster difficulties into buckets}
};

\node[box, fill=red!10, right=0.8cm of adap_filter] (output) {
  \textbf{Model Training}\\[3pt]
  $\trainedModel$\\[1pt]
  {\footnotesize trained on $\{\mathcal{C}_1, \cdots,\ \mathcal{C}_K\}$}
};

\node[box, fill=purple!12, above=1.1cm of output] (scheduler) {
  \textbf{Training Scheduler}\\[3pt]
  $\{\mathcal{C}_1, \cdots,\ \mathcal{C}_K\}$\\[1pt]
  {\footnotesize easy\;$\to$\;hard}
};


\node[box, fill=gray!10, above=1.2cm of encoder] (data) {
  \textbf{Dataset}\\[3pt]
  $\mathcal{D}=\{\bX_n\}_{n=1}^{N}$
};


\node[
  draw=blue!40, rounded corners=3pt, dashed, fill=none,
  below=0.65cm of diff,
  font=\scriptsize, align=center, text width=3.9cm, inner sep=4pt
] (strategies) {
  \textbf{Assessment Strategies}\\[2pt]
	Random, Loss, Density;\\[1pt]
	Dual-criterion: Loss+Density
};


\draw[pipearrow] (encoder) -- node[sublbl, above] {$\reprModel(\bX_n)$} (diff);

\draw[pipearrow] (diff) -- node[sublbl, above] {$\delta_n$} (adap_filter);

\draw[pipearrow]
  (adap_filter.north)
  |- (scheduler.west);

\draw[pipearrow] (scheduler.south) -- (output.north);

\draw[pipearrow]
  (data.south)
  -- (encoder.north);


\draw[->, dashed, >=Stealth, blue!50, line width=0.8pt]
  (strategies.north) -- (diff.south);


\begin{pgfonlayer}{background}
  \node[
    draw=blue!35, rounded corners=6pt, thick,
    fill=none,
    fit=(encoder)(diff)(adap_filter)(strategies),
    inner sep=10pt,
    label={[font=\small\bfseries, text=blue!60, anchor=south west]
           south west:\textbf{DCCL}},
  ] {};
\end{pgfonlayer}

\end{tikzpicture}%
}
\caption{\textbf{Schema of the proposed modular DCCL framework}.
A representation model $\reprModel$ maps each data instance $\bX_n\in\mathcal{D}$ to a vector  $\reprModel(\bX_n)\in\mathbb{R}^d$. Based on these representations, a \emph{difficulty module}~$\delta$ (dashed block) assigns a score to each instance. These scores are then partitioned into $K$ ordered curriculum buckets $\mathcal{B}_1\preceq\cdots\preceq\mathcal{B}_K$ (easy to hard) via \emph{Adaptive Filtering}. Next, the model $\trainedModel$ is fine-tuned sequentially on each training set $\mathcal{C}_k$, given the \emph{Training Scheduler}.}
\label{fig:cl_framework}
\end{figure}
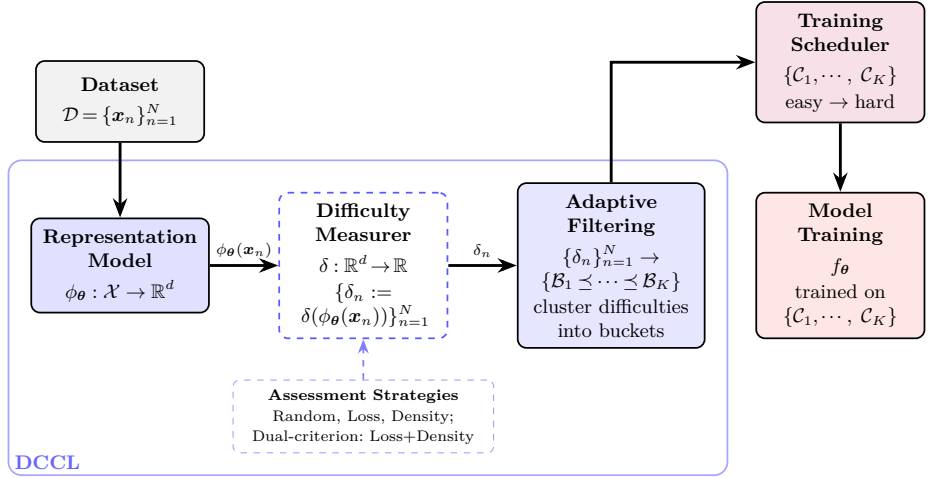

The DCCL framework is illustrated in Figure~\ref{fig:cl_framework}. The training dataset $\mathcal{D}$ is fed to a module that learns a representation
$\reprModel$ mapping each instance $\bX_n \in \mathcal{X}$ to a $d$-dimensional vector $\reprModel(\bX_n) \in \mathbb{R}^d$. The representation model $\reprModel$ can be any model that can produce a data embedding; a model of the same or a similar type as the trained model $\trainedModel$, a simple feature extractor (\eg STL decomposition in~\cite{koeneckeCurriculumLearningDeep2019}), or a pretrained model (\eg Inceptionv2 in~\cite{guoCurriculumNetWeaklySupervised2018}) on a different task or dataset than the target dataset.
Next, the embeddings obtained in the representation space $\{\reprModel(\bX_n)\}_{n=1}^N$ are evaluated by the \emph{difficulty measurer} $\delta$, which assigns a scalar difficulty score $\delta(\reprModel(\bX_n))$ to each instance.
The difficulty measurer $\delta$ can be considered as a generic and adaptable way of evaluating learning complexity: any computable scoring of the embeddings can be substituted without altering the rest of the pipeline, which ensures modularity.

These scores are then partitioned into $K$ ordered buckets $\mathcal{B}_1\preceq\cdots\preceq\mathcal{B}_K$ via an \emph{adaptive filtering} step. The model $\trainedModel$ is then trained sequentially on each training set $\mathcal{C}_k$, given the \emph{Training Scheduler}. The training scheduler is a separate design choice to the difficulty measurer that defines the learning progression of $\trainedModel$. In the rest of the work, it is kept fixed to standard strategies (One-Pass and Baby-Steps) to isolate the effect of different difficulty measures on the performance of the curriculum learning framework. In the rest of this section, we present different strategies for the difficulty measurer $\delta$, including density-based, loss-based, and hybrid approaches that rely on these two criteria. They are summarized in \Tab{tab:cl_strategies}.

\begin{figure}[t]
\centering
\includegraphics[width=0.8\linewidth]{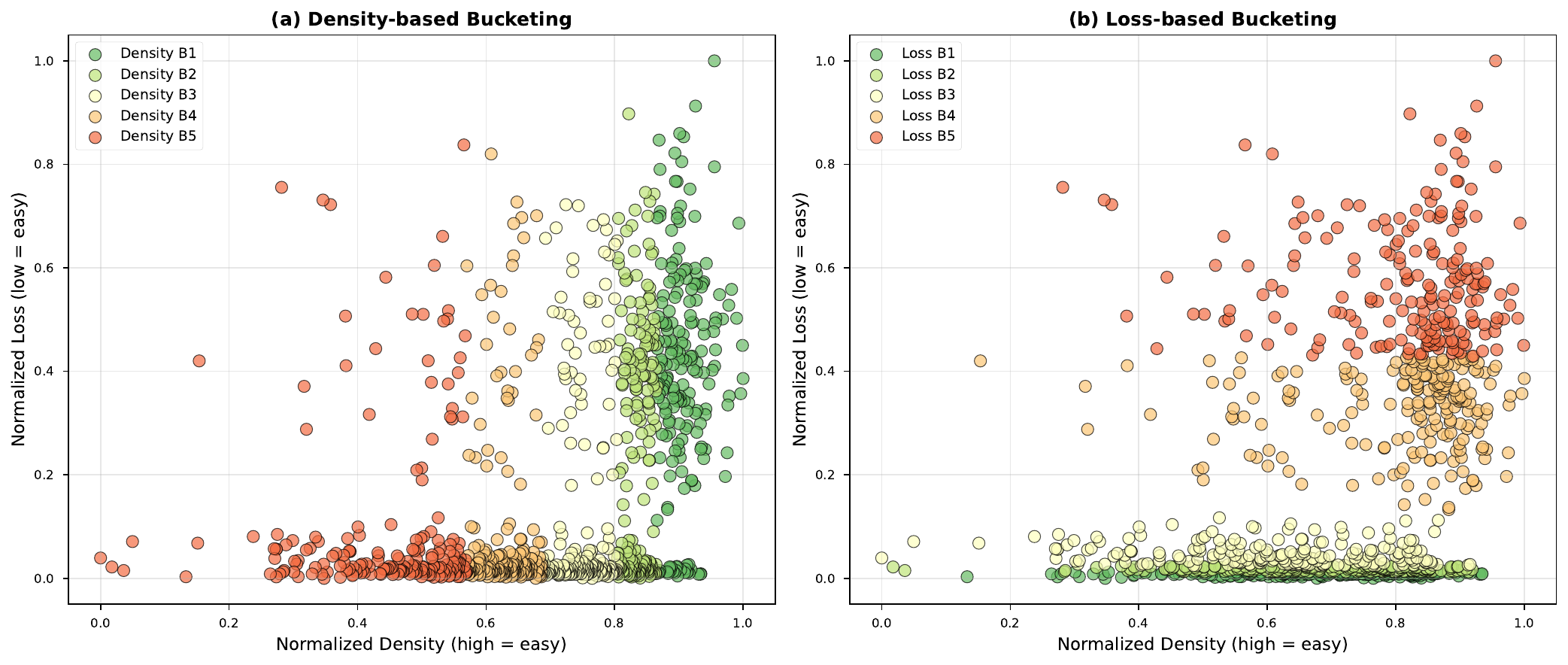}
\caption{\textbf{Scatter plot of loss vs. density scores for training instances}. The complementarity of these two difficulty criteria motivates hybrid strategies that combine both views.
}
\label{fig:loss_density_scatter}
\end{figure}

\subsection{Single-Criterion Curriculum Learning}

\inlinetitle{Loss-based Curriculum Learning}{.}
Loss-based curricula define difficulty directly from model performance~\cite{kumarSelfPacedLearningLatent2010,saxenaDataParametersNew2019}, and are therefore directly related to the learning difficulty in the context of a task. Given the target model at an initial stage, or another reference model, each instance is assigned with a score based on its associated prediction error:
\begin{align}\label{eq:delta_loss}
  \delta_{\text{loss}}(\bX_n) = \ell\!\left(\reprModel(\bX_n),\, \by_n\right),
\end{align}
where $\ell$ is the training loss.
Instances with lower loss are considered easier, while high-loss instances are treated as harder. The resulting scores are ranked and partitioned into curriculum buckets of increasing difficulty. This strategy is adaptive to the optimization dynamics and often emphasizes instances that are currently poorly fitted by the model. In practice, it can be implemented either from a warm-up model trained on all data, or by periodically refreshing the scores during curriculum training.

\inlinetitle{Density-based Curriculum Learning}{.}
Compared to loss-based CL, density-based CL has been studied to a much lesser extent, mostly for classification tasks, where the data density is used to retrieve the least noisy data instances (\ie the \emph{cleanest} labels) for training~\cite{guoCurriculumNetWeaklySupervised2018,chaudhryDataDistributionBasedCurriculum2024}.
This principle can be extended to the setting where no labels are used in the density estimation step: instances located in dense regions are treated as easier (more prototypical), while instances in sparse regions are treated as harder (rarer or more atypical). Therefore, sorting by this score induces an easy-to-hard curriculum without using labels.

Contrary to loss-based criteria, pure density-based criteria are less directly connected to the learning difficulty. However, the proposed DCCL framework employs a supervised representation learning step that can mitigate this gap. The idea supporting this option is that a powerful model would bring data instances with similar properties (\eg same label) closer in the embedding space. The existing wealth of powerful such models for a wide range of data types, including time-series or other structured data like graphs, make the density-based measures feasible to utilize, and thus render our proposed framework of generic interest. On the downside, the learned data embedding may be misaligned with how the target model we want to eventually learn sees the data, hence the computed data density may not translate well to gains in the performance of the latter.

In the following, we present two approaches to density estimation in the representation space $\reprModel:\mathcal{X}\to\mathbb{R}^d$.
$k$-Nearest Neighbors Density ($k$-NN) is inspired by~\cite{guoCurriculumNetWeaklySupervised2018}, and assigns to each instance a local density score by assessing density at the $k$-NN level. For each instance $\bX_n$, its density $\rho_{k\text{NN}}(\reprModel(\bX_n))$ is estimated by counting its $k$-NNs within a fixed radius in the embedded space.
The second one is Kernel Density Estimation (KDE)~\cite{weglarczykKernelDensityEstimation2018}, which estimates a smooth probability density by summing kernel contributions around each embedded instance. For a given instance $\bX_n$, the density is computed as $\hat{p}_{\text{KDE}}(\bX_n)=\frac{1}{N}\sum_i K_h(\reprModel(\bX_n)-\reprModel(\bX_i))$, where $K_h$ is a kernel function with bandwidth $h$.
Using these estimates as difficulty scores provides a continuous density-based ordering less sensitive to discrete neighborhood boundaries.

\subsection{Dual-Criterion Curriculum Learning}

Density and loss-based views provide complementary criteria: density captures intrinsic structure in representation space, while loss reflects current model fit, cf. Figure~\ref{fig:loss_density_scatter}. Passing from a single criterion to using multiple criteria for determining a curriculum means that a way to fuse those criteria needs to be defined.
We explore three hybrid variants; the first two use a convex combination of the two criteria to produce a single measurement, while the third one uses $2$D buckets.

\begin{figure}[t]
\centering
\includegraphics[width=0.8\linewidth]{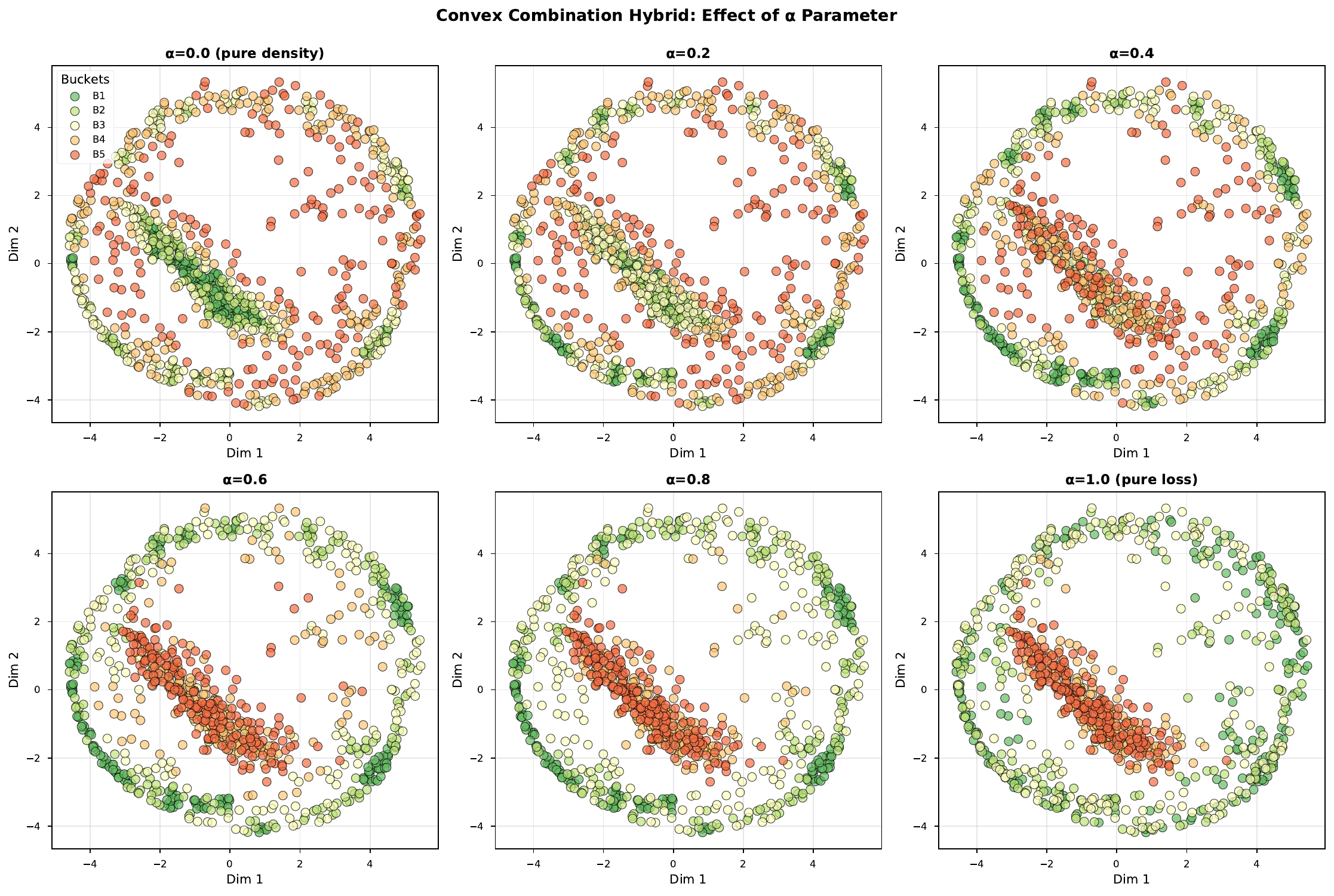}
\caption{\textbf{Effect of the convex-mixing parameter $\alpha$ on curriculum buckets}. As $\alpha$ varies from $0$ (pure density-based) to $1$ (pure loss-based), the resulting difficulty ordering smoothly interpolates between the strategies.}
\label{fig:hybrid_convex}
\end{figure}

\inlinetitle{Convex-value hybrid}{.}~%
For each instance $\bX_n$, we compute two $[0,1]$-normalized scores: the loss-based $\tilde{\delta}_{\text{loss}}(\bX_n)$ (Equation~\ref{eq:delta_loss}) and density-based $\tilde{\delta}_{\text{dens}}(\bX_n)$ (high density is interpreted as easiness). We then define the hybrid difficulty score:
\[
  \delta_{\alpha\text{-val}}(\bX_n)
  = \alpha\,\tilde{\delta}_{\text{loss}}(\bX_n)
  + (1-\alpha)\,\tilde{\delta}_{\text{dens}}(\bX_n),
  \qquad \alpha,\ \tilde{\delta}_{\text{loss}}(\bX_n),\ \tilde{\delta}_{\text{dens}}(\bX_n) \in [0,1].
\]
Instances are sorted by $\delta_{\alpha\text{-val}}$ (easy\,$\rightarrow$\,hard) and split into $K$ buckets. The learnable hyperparameter $\alpha$ controls the trade-off: $\alpha=1$ gives a purely loss-based curriculum, $\alpha=0$ a purely density-based one, as shown in Figure~\ref{fig:hybrid_convex}.

\inlinetitle{Convex-rank hybrid}{.}~%
Instead of combining normalized values, this variant combines \emph{ranks}. Let $r_{\text{loss}}(\bX_n)$ be the rank induced by loss, and $r_{\text{dens}}(\bX_n)$ the rank induced by density. The fused score is now defined as:
\[
  \delta_{\alpha\text{-rank}}(\bX_n) = \alpha\,r_{\text{loss}}(\bX_n) + (1-\alpha)\,r_{\text{dens}}(\bX_n).
  \qquad \alpha\in[0,1],
\]
Instances are then sorted in increasing $\delta_{\alpha\text{-rank}}$ and split into $K$ buckets. As before, $\alpha=0$ yields a purely density-based curriculum and $\alpha=1$ a purely loss-based one.

\begin{figure}[t]
\centering
\includegraphics[width=0.8\linewidth]{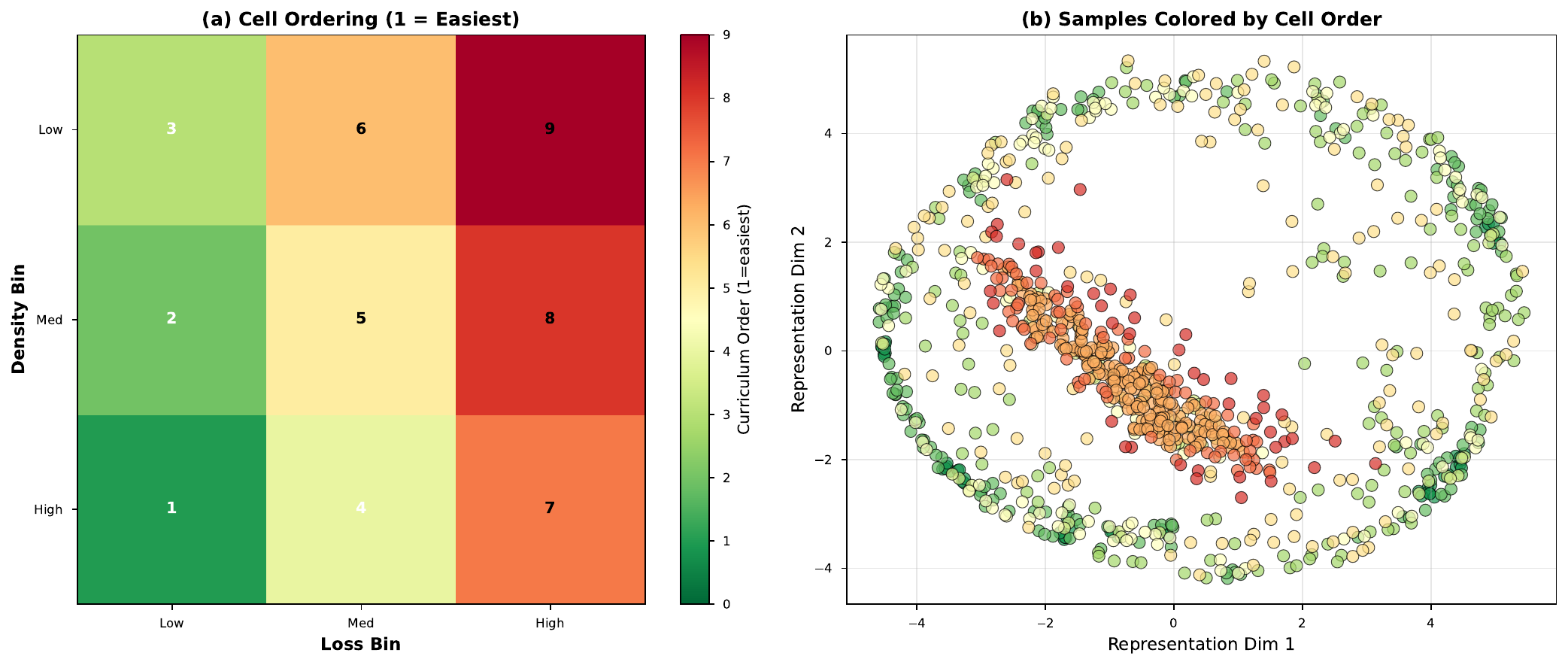}
\caption{\textbf{Bivariate loss-density stratification.} Instances are binned into a 2D grid according to their loss and density scores, then ordered cell-by-cell from easy (low loss, high density) to hard (high loss, low density). Each cell is taken as a bucket.}
\label{fig:hybrid_bivariate}
\end{figure}

\inlinetitle{Grid loss-density stratification (Grid)}{.}~%
We discretize independently the loss and density scores into bins (\eg $K$ quantile bins), forming a
2D stratification of cells over the \{low/high loss\} $\times$ \{high/low density\}
plane, see Figure~\ref{fig:hybrid_bivariate}. Cells are then ordered from easy (low loss, high density) to hard (low density, high loss), and merged/split if needed to produce exactly $K$ curriculum buckets. This version preserves a structured view of the joint loss-density landscape and can better capture heterogeneous regimes compared to a single scalar score.

\begin{table}[t]
\centering\small
\caption{\textbf{Summary of the CL strategies.} Representation model $\reprModel$ and difficulty measurer $\delta$. High $\delta(\bX_n)$ suggests a harder instance; $\rho$ is the local neighbourhood density, $\hat{p}$ is the kernel density estimate, and $\ell$ is the training loss.}
\label{tab:cl_strategies}
\resizebox{\textwidth}{!}{%
\begin{tabular}{p{2cm}p{2cm}p{4.5cm}p{6cm}}
\toprule
\textbf{Strategy} & \textbf{View} & \textbf{Repr.\ model $\reprModel$} & \textbf{Difficulty $\delta(\bX_n)$} \\
\midrule
STL~\cite{koeneckeCurriculumLearningDeep2019}
  & STL residuals
  & $\reprModel(\bX_n) = \mathbf{T}_n+\mathbf{S}_n+\mathbf{R}_n$
  & Magnitude of residuals $||\mathbf{R}_n||$ \\
Loss-based
  & Loss
  & Forecasting $\reprModel:\mathcal{X}\to\mathcal{Y}$
  & $\ell\!\left(\reprModel(\bX_n),\,\by_n\right)$ \\
$k$-NN
  & Density
  & Encoder $\reprModel:\mathcal{X}\to\mathbb{R}^{d}$
  & $-\rho\!\left(\reprModel(\bX_n)\right)$ \\
KDE
  & Density
  & Encoder $\reprModel:\mathcal{X}\to\mathbb{R}^{d}$
  & $-\hat{p}\!\left(\reprModel(\bX_n)\right)$ \\
Convex-value
  & Loss+Density
  & Encoder $+$ forecasting model
  & $\alpha\,\tilde{\delta}_{\mathrm{loss}}(\bX_n)+(1{-}\alpha)\,\tilde{\delta}_{\mathrm{dens}}(\bX_n),\quad\alpha\in[0,1]$ \\
Convex-ranks
  & Loss+Density
  & Encoder $+$ forecasting model
  & $\alpha\,r_{\mathrm{loss}}(\bX_n)+(1{-}\alpha)\,r_{\mathrm{dens}}(\bX_n),\quad\alpha\in[0,1]$ \\
Grid
  & Loss+Density
  & Encoder $+$ forecasting model
  & Cell ordering on quantile grid $\{\delta_{\mathrm{loss}}\times\delta_{\mathrm{dens}}\}$ \\
\bottomrule
\end{tabular}%
}
\end{table}

\section{Experiments}

This section evaluates the proposed framework on benchmark time-series datasets under One-Pass and Baby-Steps schedules. We report dataset and training configurations, then present quantitative comparisons across curriculum strategies using mean and standard deviation over multiple runs.

\subsection{Datasets}
We evaluate our CL framework on \numberdatasets widely adopted multivariate time-series forecasting benchmarks, spanning diverse application domains (energy, transportation, weather, and public health) and temporal resolutions.

\begin{itemize}[leftmargin=1.2em,itemsep=.4em,topsep=.4em]
  \item \textbf{Electricity.} A dataset including hourly electricity consumption data from $2016$ to $2019$ for $320$ clients. Each time-series corresponds to a single client's power usage. We use as target the sum of all clients' consumption.
  \item \textbf{ETT.} A benchmark dataset with hourly data recorded from July 2016 to July 2018. It includes seven features: one target variable (oil temperature) and six power load covariates measured from electricity transformers.

  \item \textbf{Weather.} It includes $21$ weather indicators, such as air temperature and humidity. Its data is recorded every $10$ minutes during $2020$ in Germany.
  \item \textbf{ILI.} The ILI dataset tracks weekly ratios of patients diagnosed with influenza-like illness relative to total outpatient visits across US clinical facilities from $2002$ to $2021$. The time-series displays strong annual seasonality with occasional regime shifts due to epidemic events.
  \item \textbf{Solar AL.} This dataset contains the solar power production records in the year of $2006$ in Alabama, which is sampled every $10$ minutes from $137$ PV plants in Alabama State. We use as a target the total solar power production of all plants.
\end{itemize}

\subsection{Experimental settings}
Following established benchmarking practices, we 
perform sliding-window forecasting with dataset-specific lookback $L$, horizon $H$, and stride $s$. All datasets are partitioned chronologically into training, validation, and test splits with ratios $70\%/15\%/15\%$ to ensure temporal consistency and prevent data leakage. Normalization statistics are computed on the training split only and then applied to validation and test. For optimization and curriculum training, we use Adam; mini-batch sizes are dataset-dependent and are specified in Table~\ref{tab:dataset_settings}.

\inlinetitle{Comparison with CL strategies}{.}
To showcase the proposed CL framework (Section~\ref{sec:CL_TS}), we compare to the following learning strategies:
\begin{itemize}
  \item No-curriculum: one-shot model training with no data bucketing.
  \item Random: random buckets of training data are selected.
  \item STL-based CL (from~\cite{koeneckeCurriculumLearningDeep2019}): Instances are processed with the Seasonal Trend decomposition using loss~\cite{cleveland1990stl}. The residuals obtained from this decomposition are used as a difficulty score.

\end{itemize}

\begin{table}[t]
\centering\small
\caption{Dataset-specific forecasting and sampling settings used in experiments.}
\label{tab:dataset_settings}
\resizebox{\textwidth}{!}{%
\begin{tabular}{p{2.5cm}p{1cm}p{1cm}p{1cm}ccc}
\toprule
\textbf{Dataset} & \textbf{$L$} & \textbf{$H$} & \textbf{$s$} & \textbf{Sample Rate} &\textbf{Subsample} & \textbf{Batch size (repr/curr)} \\
\midrule
Electricity           & 96 & 24 & 1 & 1\text{h} & 2000 & 32 \\
ETT           & 96 & 24 & 1 & 1\text{h} & 2000 & 8 / 32 \\
Weather       & 12  & 4 & 6 & 10 \text{min}& 500 & 8 / 32 \\
ILI           & 24  & 4 & 1 &1\text{week} & None
& 8 / 32 \\
Solar AL         & 96  & 24 & 6 & 10 \text{min} & 2000 & 8 / 32 \\
\bottomrule
\end{tabular}
}
\end{table}

\inlinetitle{Training and Optimization Setup}{.}~%
Our implementation follows a two-stage procedure: (i) representation 
learning and difficulty estimation, and (ii) curriculum model training.

\paragraph{Stage 1: representation model training and difficulty extraction.}
We train a representation model on the full training set (with shuffled mini-batches), using MSE loss for forecasting and Adam with gradient clipping. Hyperparameters are selected by optimization (see below), with search ranges including hidden dimension in $\{32,64,128\}$ and learning rate in $[10^{-4},5\times 10^{-3}]$ (log scale).

\paragraph{Representation architectures.}
For each training instance, we build a single embedding with one of two sequence encoders. With the LSTM forecaster, the representation is the final cell state $\mathbf{c}_{L}$ after reading the full lookback window,
$\reprModel(\bX_n)=\mathbf{c}_{L}$. With the Transformer forecaster, we use the last-layer ($\ell_{\max}$) hidden state at the final time index,
$\reprModel(\bX_n)=\mathbf{h}^{(\ell_{\max})}_{L}$. In both cases, embeddings are extracted over the entire training set with a deterministic loader, ensuring one-to-one alignment between representation indices and dataset indices for downstream bucketing.
For loss-based and hybrid strategies, per-instance losses are also computed from the same trained representation model using unreduced MSE averaged per instance.

\paragraph{Stage 2: curriculum model training.}
Given ordered buckets, we train a forecasting curriculum model (Transformer in our default setup) for 10 epochs per curriculum stage, with Adam and mini-batch size 32. We use $K=5$ buckets and evaluate two schedules: One-Pass and Baby-Steps; by default, experiments use One-Pass. These two schedules are selected as they are the most commonly used in the curriculum learning literature, and they represent two distinct paradigms. On the one hand, One-Pass emphasizes a strict progression from easy to hard: exhibiting the ``raw'' effect of difficulty measures. Baby-Steps, on the other hand, allows for revisiting easier instances, which can be beneficial for stability and generalization. Evaluating both schedules provides a comprehensive view of how curriculum strategies perform under different training dynamics.

\inlinetitle{Validation protocol and hyperparameter optimization}{.}~%
The DCCL pipeline has hyperparameters at multiple levels (representation model, difficulty measurer, adaptive filtering/bucketing, scheduler, and curriculum forecaster). We separate them into fixed settings and tuned settings as follows:
\begin{itemize}[leftmargin=1.2em,itemsep=.25em,topsep=.25em]
  \item \textbf{Fixed across trials.} Data splits, optimizer family (Adam), gradient clipping (max norm $1.0$), number of buckets ($K=5$), and stage budget (10 epochs per stage). The schedule is treated as an experimental factor (One-Pass or Baby-Steps), not as a free parameter inside a trial.
  \item \textbf{Optimized by validation.} Representation-model hyperparameters (\eg hidden dimension and learning rate), strategy-specific difficulty-measurer hyperparameters (\eg density estimator choice and its internal parameters), and dual-criterion fusion parameters (\eg $\alpha$ and $2$D grid binning settings).
\end{itemize}
The tuned parameters are selected with Optuna (TPE sampler, seed 42, 30 trials). In each trial, we: (1) train the representation model, (2) build buckets from extracted states/losses according to the selected strategy, (3) train the curriculum model with the selected schedule, and (4) rank the trial by validation MSE.

\paragraph{Final evaluation.}
After selecting the best configuration, we rerun curriculum training for $10$ independent seeds and report aggregate statistics (mean and standard deviation) on validation and test MSE, along with training time.

\subsection{Experimental Results}

\begin{table*}[t]
\centering
\caption{\textbf{Test losses (MSE $\downarrow$) across curriculum strategies and datasets}. Values are mean $\pm$ std over $10$ runs. \textbf{Bold} and \underline{underlined} entries denote best and second-best performance \emph{within each dataset and schedule} (One-Pass or Baby-Steps). The last column reports the mean rank ($\downarrow$) per strategy over datasets (ties averaged). Weather${}^\dagger$ values are multiplied by $10^{2}$ for readability.}
\label{tab:test_losses_merged}
\resizebox{\textwidth}{!}{%
\begin{tabular}{ll|ccccc|c}
\toprule
& & \multicolumn{5}{c|}{\textbf{Dataset}} & \textbf{Mean rank} \\
 & \textbf{Strategy}
& \textbf{Electricity}
& \textbf{ETT}
& \textbf{ILI}
& \textbf{Weather}${}^\dagger$
& \textbf{Solar AL}
& $\downarrow$ \\
\midrule
\multirow{9}{*}{\rotatebox[origin=c]{90}{\textbf{One-Pass}}}
& No-curriculum      & \shadeOP{0.852}{0.486 $\pm$ 0.065} & \shadeOP{0.948}{1.005 $\pm$ 0.085} & \shadeOP{0.828}{0.865 $\pm$ 0.019} & \shadeOP{0.900}{0.201 $\pm$ 0.036} & \shadeOP{0.852}{0.080 $\pm$ 0.009} & \shadeOP{0.876}{5.00} \\
& Random             & \shadeOP{0.828}{0.462 $\pm$ 0.059} & \shadeOP{0.924}{1.004 $\pm$ 0.159} & \shadeOP{0.852}{0.871 $\pm$ 0.043} & \shadeOP{0.876}{0.196 $\pm$ 0.060} & \shadeOP{0.828}{0.076 $\pm$ 0.007} & \shadeOP{0.852}{4.40} \\
& STL & \shadeOP{0.970}{0.604 $\pm$ 0.069} & \shadeOP{0.900}{0.928 $\pm$ 0.137} & \shadeOP{0.900}{0.932 $\pm$ 0.031} & \shadeOP{0.948}{0.226 $\pm$ 0.071} & \shadeOP{0.876}{0.082 $\pm$ 0.007} & \shadeOP{0.970}{6.80} \\
\cmidrule{2-8}
& Loss-based         & \shadeOP{0.780}{\textbf{0.399 $\pm$ 0.026}} & \shadeOP{0.970}{1.051 $\pm$ 0.627} & \shadeOP{0.948}{0.946 $\pm$ 0.040} & \shadeOP{0.816}{\underline{0.169 $\pm$ 0.025}} & \shadeOP{0.900}{0.103 $\pm$ 0.006} & \shadeOP{0.900}{5.30} \\

\cmidrule{2-8}
& $k$-NN             & \shadeOP{0.948}{0.532 $\pm$ 0.101} & \shadeOP{0.852}{0.582 $\pm$ 0.159} & \shadeOP{0.970}{1.092 $\pm$ 0.191} & \shadeOP{0.816}{\underline{0.169 $\pm$ 0.053}} & \shadeOP{0.970}{0.186 $\pm$ 0.022} & \shadeOP{0.948}{6.50} \\
& KDE                & \shadeOP{0.804}{\underline{0.435 $\pm$ 0.042}} & \shadeOP{0.876}{0.830 $\pm$ 0.106} & \shadeOP{0.876}{0.925 $\pm$ 0.034} & \shadeOP{0.924}{0.209 $\pm$ 0.023} & \shadeOP{0.948}{0.115 $\pm$ 0.024} & \shadeOP{0.924}{5.40} \\
\cmidrule{2-8}
& Convex-value       & \shadeOP{0.876}{0.502 $\pm$ 0.074} & \shadeOP{0.780}{\textbf{0.129 $\pm$ 0.015}} & \shadeOP{0.804}{\underline{0.862 $\pm$ 0.029}} & \shadeOP{0.970}{0.470 $\pm$ 0.047} & \shadeOP{0.804}{\underline{0.074 $\pm$ 0.008}} & \shadeOP{0.804}{\underline{3.80}} \\
& Convex-ranks       & \shadeOP{0.924}{0.513 $\pm$ 0.025} & \shadeOP{0.804}{\underline{0.267 $\pm$ 0.188}} & \shadeOP{0.780}{\textbf{0.821 $\pm$ 0.021}} & \shadeOP{0.852}{0.177 $\pm$ 0.026} & \shadeOP{0.924}{0.106 $\pm$ 0.018} & \shadeOP{0.828}{4.20} \\
& Grid               & \shadeOP{0.900}{0.505 $\pm$ 0.039} & \shadeOP{0.828}{0.375 $\pm$ 0.108} & \shadeOP{0.924}{0.941 $\pm$ 0.029} & \shadeOP{0.780}{\textbf{0.148 $\pm$ 0.023}} & \shadeOP{0.780}{\textbf{0.069 $\pm$ 0.005}} & \shadeOP{0.780}{\textbf{3.60}} \\
\midrule\midrule
\multirow{9}{*}{\rotatebox[origin=c]{90}{\textbf{Baby-Steps}}}
& No-curriculum      & \shadeBS{0.876}{0.486 $\pm$ 0.065} & \shadeBS{0.970}{1.005 $\pm$ 0.085} & \shadeBS{0.828}{0.865 $\pm$ 0.019} & \shadeBS{0.970}{0.201 $\pm$ 0.036} & \shadeBS{0.970}{0.080 $\pm$ 0.009} & \shadeBS{0.970}{7.00} \\
& Random             & \shadeBS{0.900}{0.518 $\pm$ 0.070} & \shadeBS{0.900}{0.935 $\pm$ 0.123} & \shadeBS{0.852}{0.805 $\pm$ 0.043} & \shadeBS{0.792}{\textbf{0.026 $\pm$ 0.014}} & \shadeBS{0.876}{0.063 $\pm$ 0.005} & \shadeBS{0.852}{4.30} \\
& STL & \shadeBS{0.970}{0.630 $\pm$ 0.098} & \shadeBS{0.876}{0.914 $\pm$ 0.109} & \shadeBS{0.924}{0.882 $\pm$ 0.021} & \shadeBS{0.900}{0.031 $\pm$ 0.016} & \shadeBS{0.828}{\underline{0.060 $\pm$ 0.007}} & \shadeBS{0.912}{6.00} \\
\cmidrule{2-8}
& Loss-based         & \shadeBS{0.924}{0.532 $\pm$ 0.044} & \shadeBS{0.948}{0.987 $\pm$ 0.132} & \shadeBS{0.876}{0.831 $\pm$ 0.026} & \shadeBS{0.948}{0.077 $\pm$ 0.069} & \shadeBS{0.900}{0.066 $\pm$ 0.012} & \shadeBS{0.948}{6.60} \\
\cmidrule{2-8}
& $k$-NN             & \shadeBS{0.852}{0.507 $\pm$ 0.035} & \shadeBS{0.804}{\underline{0.556 $\pm$ 0.135}} & \shadeBS{0.970}{0.952 $\pm$ 0.110} & \shadeBS{0.792}{\textbf{0.026 $\pm$ 0.016}} & \shadeBS{0.948}{0.070 $\pm$ 0.016} & \shadeBS{0.876}{5.10} \\
& KDE                & \shadeBS{0.828}{0.492 $\pm$ 0.049} & \shadeBS{0.924}{0.969 $\pm$ 0.145} & \shadeBS{0.948}{0.921 $\pm$ 0.096} & \shadeBS{0.852}{\underline{0.029 $\pm$ 0.008}} & \shadeBS{0.924}{0.069 $\pm$ 0.016} & \shadeBS{0.912}{6.00} \\
\cmidrule{2-8}
& Convex-value       & \shadeBS{0.804}{\underline{0.483 $\pm$ 0.047}} & \shadeBS{0.780}{\textbf{0.298 $\pm$ 0.143}} & \shadeBS{0.900}{0.874 $\pm$ 0.104} & \shadeBS{0.924}{0.038 $\pm$ 0.018} & \shadeBS{0.828}{\underline{0.060 $\pm$ 0.007}} & \shadeBS{0.816}{\underline{3.80}} \\
& Convex-ranks       & \shadeBS{0.948}{0.548 $\pm$ 0.043} & \shadeBS{0.828}{0.757 $\pm$ 0.332} & \shadeBS{0.780}{\textbf{0.714 $\pm$ 0.018}} & \shadeBS{0.852}{\underline{0.029 $\pm$ 0.015}} & \shadeBS{0.828}{\underline{0.060 $\pm$ 0.012}} & \shadeBS{0.816}{\underline{3.80}} \\
& Grid               & \shadeBS{0.780}{\textbf{0.479 $\pm$ 0.064}} & \shadeBS{0.852}{0.847 $\pm$ 0.164} & \shadeBS{0.804}{\underline{0.749 $\pm$ 0.032}} & \shadeBS{0.852}{\underline{0.029 $\pm$ 0.016}} & \shadeBS{0.780}{\textbf{0.058 $\pm$ 0.007}} & \shadeBS{0.780}{\textbf{2.40}} \\
\bottomrule
\end{tabular}%
}
\end{table*}

\begin{figure*}[t]
\centering
\subfloat{\includegraphics[width=0.8\textwidth]{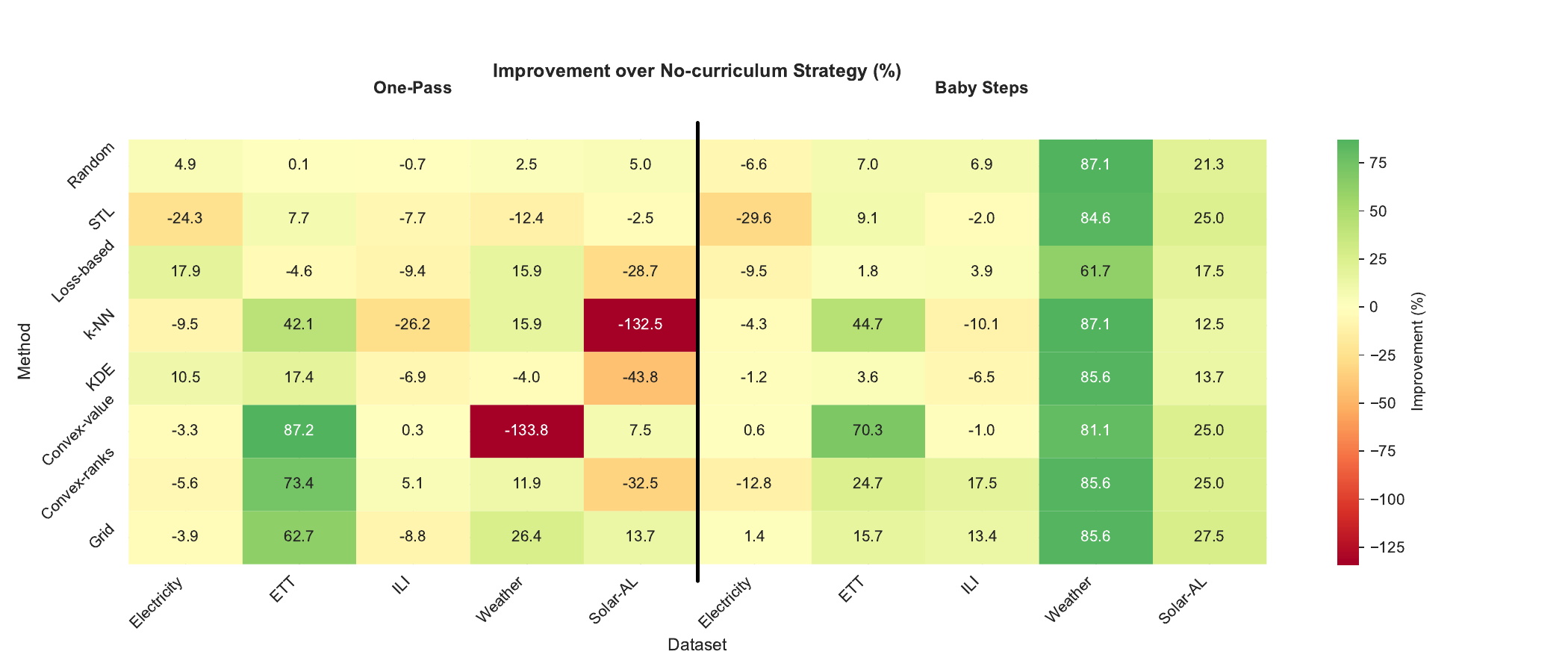}}\\
\subfloat{\includegraphics[width=0.7\textwidth]{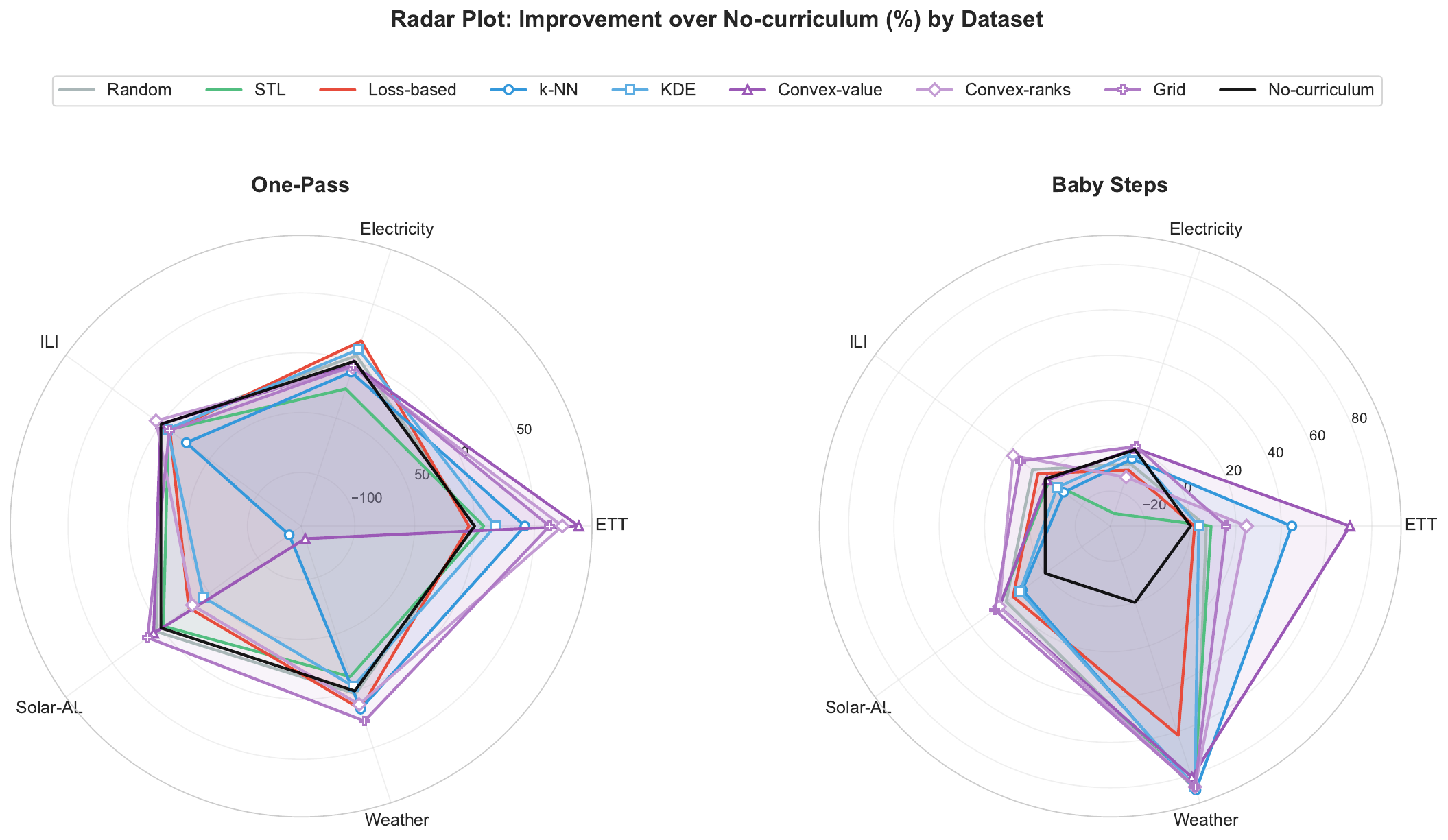}}
\caption{\textbf{Percentage improvement over the No-curriculum baseline.} Top: The results concern each strategy (rows) and dataset/schedule combination (columns). Green cells indicate improvement; red cells indicate degradation. Bottom: Radar plots of percentage improvement over No-curriculum per strategy, separately for the One-Pass (left) and Baby-Steps (right) schedules. Each axis corresponds to a dataset; polygons further from the black reference circle (0\% improvement) indicate stronger gains.}
\label{fig:improvement_heatmap}
\end{figure*}

Table~\ref{tab:test_losses_merged} reports test MSE across five datasets and both One-Pass and Baby-Steps schedules. Over these datasets, hybrid strategies achieve the best aggregate ranks: under One-Pass, Grid and Convex-value obtain mean ranks of $3.60$ and $3.80$; under Baby-Steps, Grid obtains $2.40$, followed by Convex-value and Convex-ranks.
The same trend is visible in Figure~\ref{fig:improvement_heatmap}: where hybrid rows show more consistently positive improvements over the No-curriculum baseline. The radar plots complement this view by showing schedule-specific profiles: Baby-Steps generally amplifies gains on datasets such as ILI and Weather, whereas One-Pass exhibits more heterogeneous improvements across methods.

\inlinetitle{Findings. Hybrid strategies provide the most consistent gains}{:} The results indicate that combining density and loss yields the most robust curricula across datasets and schedules. In particular, hybrid variants dominate the top mean-rank positions in both schedules, supporting the view that these two criteria capture complementary aspects of instance difficulty.
This confirms the central hypothesis of DCCL, here in the chosen application: instance difficulty in time-series forecasting has at least two complementary faces: \emph{intrinsic complexity}, captured by the position of an instance in the representation space relative to the rest of the data, and \emph{model-relative hardness}, captured by prediction error. Neither criterion alone is sufficient: loss-based curricula can be misleading when the warm-up model is not yet well calibrated, while density-based curricula are blind to the learning dynamics of the forecaster. Combining them, even with the simple convex combination, reliably bridges this gap.

\inlinetitle{One-Pass vs Baby-Steps schedules}{:}
Baby-Steps produces lower mean ranks for the top hybrid strategies (Grid: $2.40$ vs $3.60$ for One-Pass) by re-exposing the model to easy instances at each stage. This comes at a computational cost as the effective training set grows by one bucket at each stage, multiplying total gradient steps by $\tfrac{K+1}{2}$ relative to One-Pass. In practice, for datasets with small training sets (ILI, Weather), Baby-Steps is affordable and clearly beneficial; for large datasets such as Electricity, the increased cost merits consideration. Notably, for strategies that already achieve strong performance under One-Pass (\eg Convex-value on ETT), the Baby-Steps advantage is more modest.

\begin{figure}[t]
  \centering
    \hspace{-1.5em}
  \subfloat[Electricity dataset]{\includegraphics[width=0.4\textwidth]{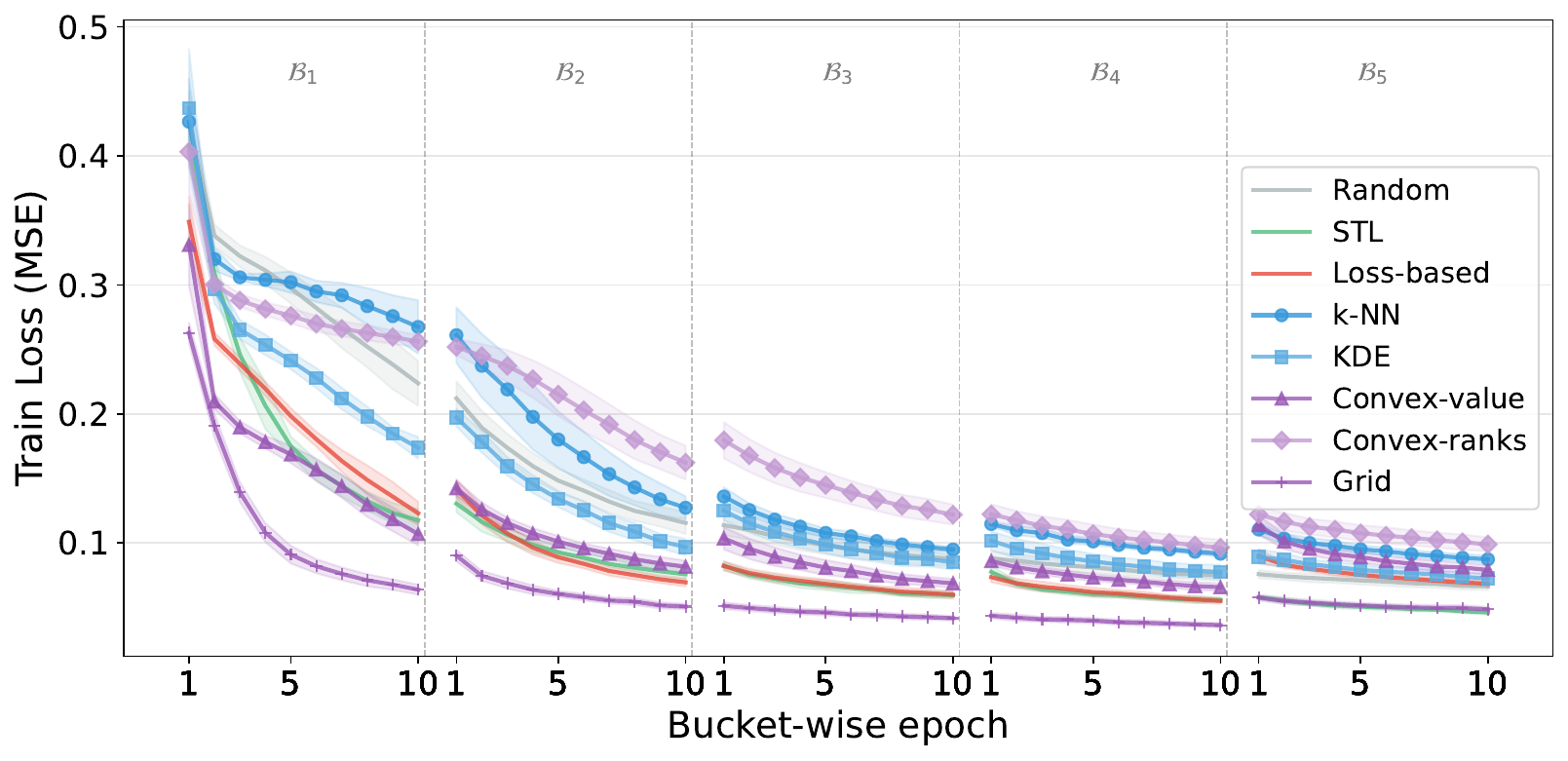}}
  \hspace{0.5em}
  \subfloat[ILI dataset]{\includegraphics[width=0.4\textwidth]{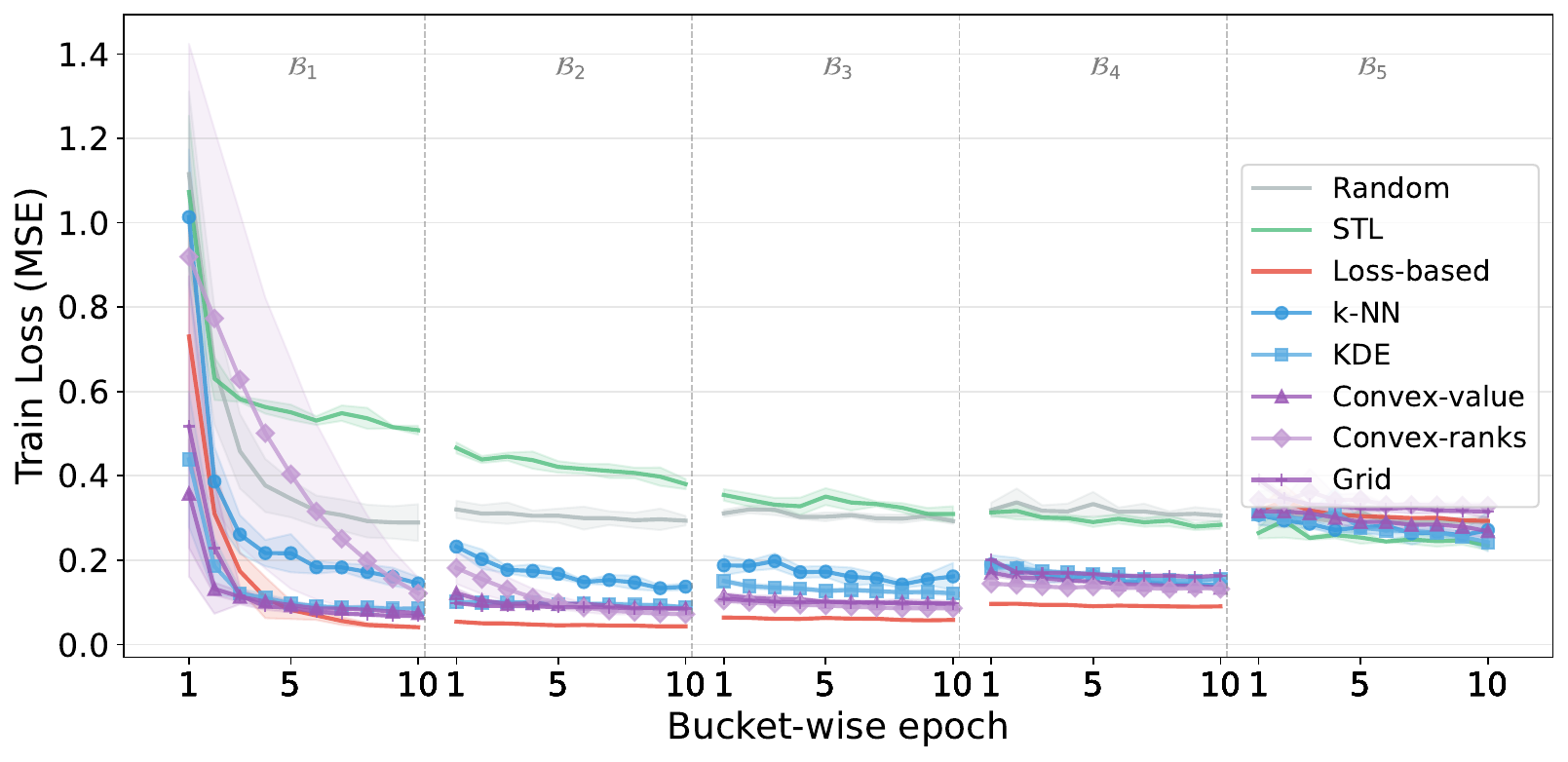}}\\
  \hspace{-3em}
  \subfloat[ETT dataset,\\\phantom{xxx}STL vs Convex-value]{\includegraphics[width=0.3\textwidth]{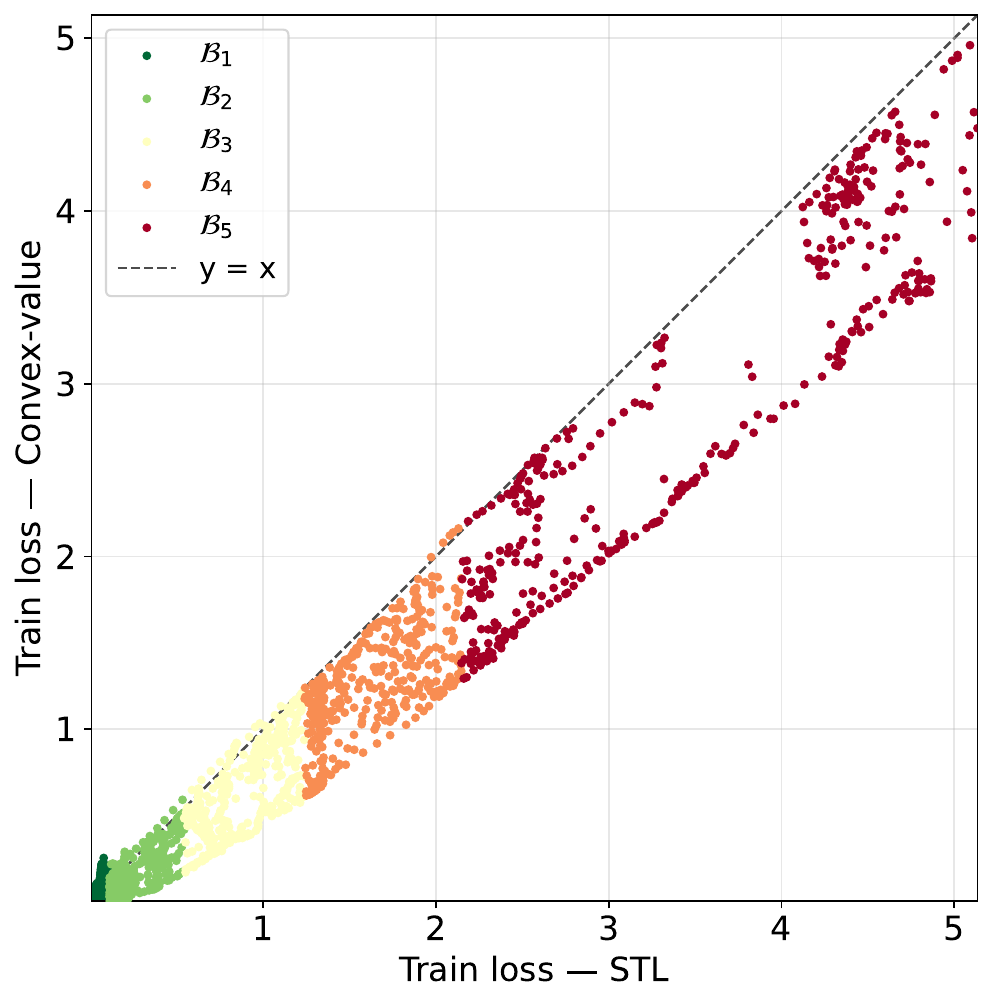}}
  \hspace{0.4em}
  \subfloat[Solar dataset,  KDE v. Loss-based]{\includegraphics[width=0.31\textwidth]{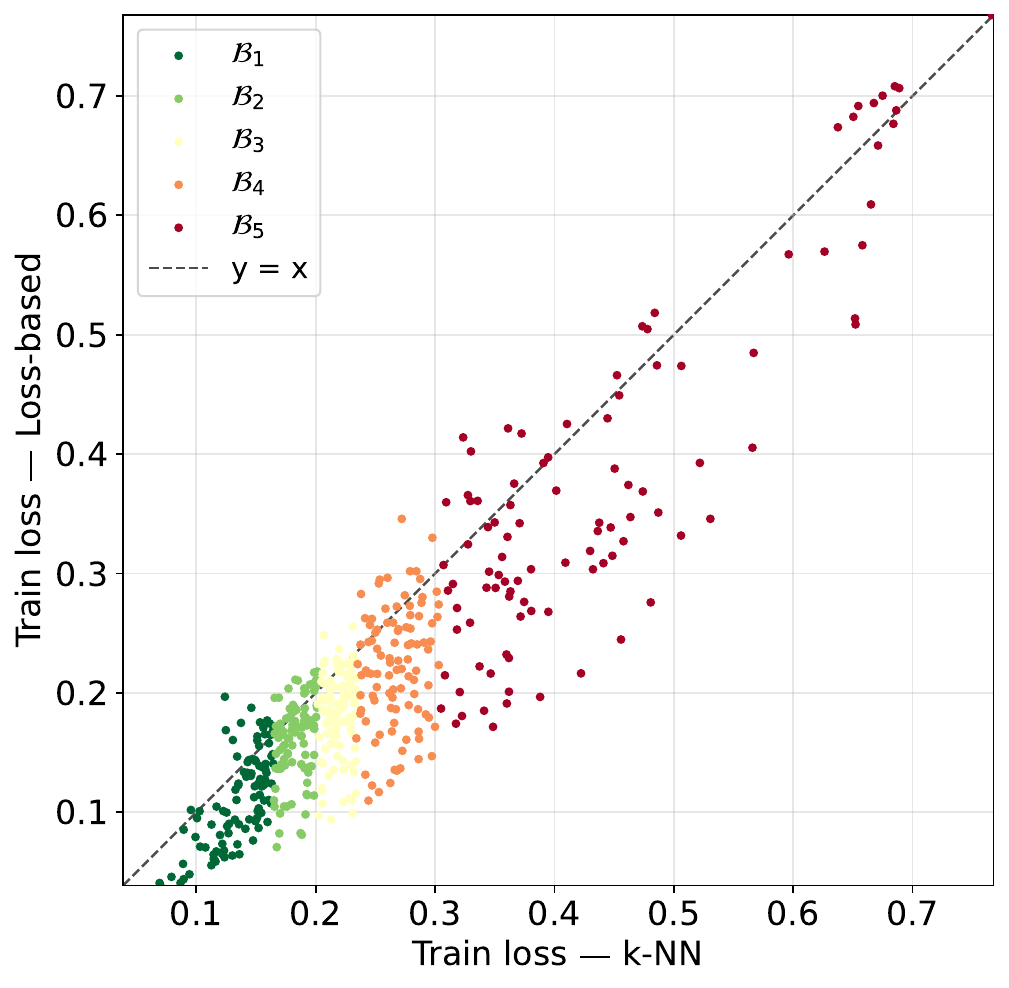}}
  \hspace{0.4em}
  \subfloat[Weather dataset,\\\phantom{xxx}KDE v. Loss-based]{\includegraphics[width=0.325\textwidth]{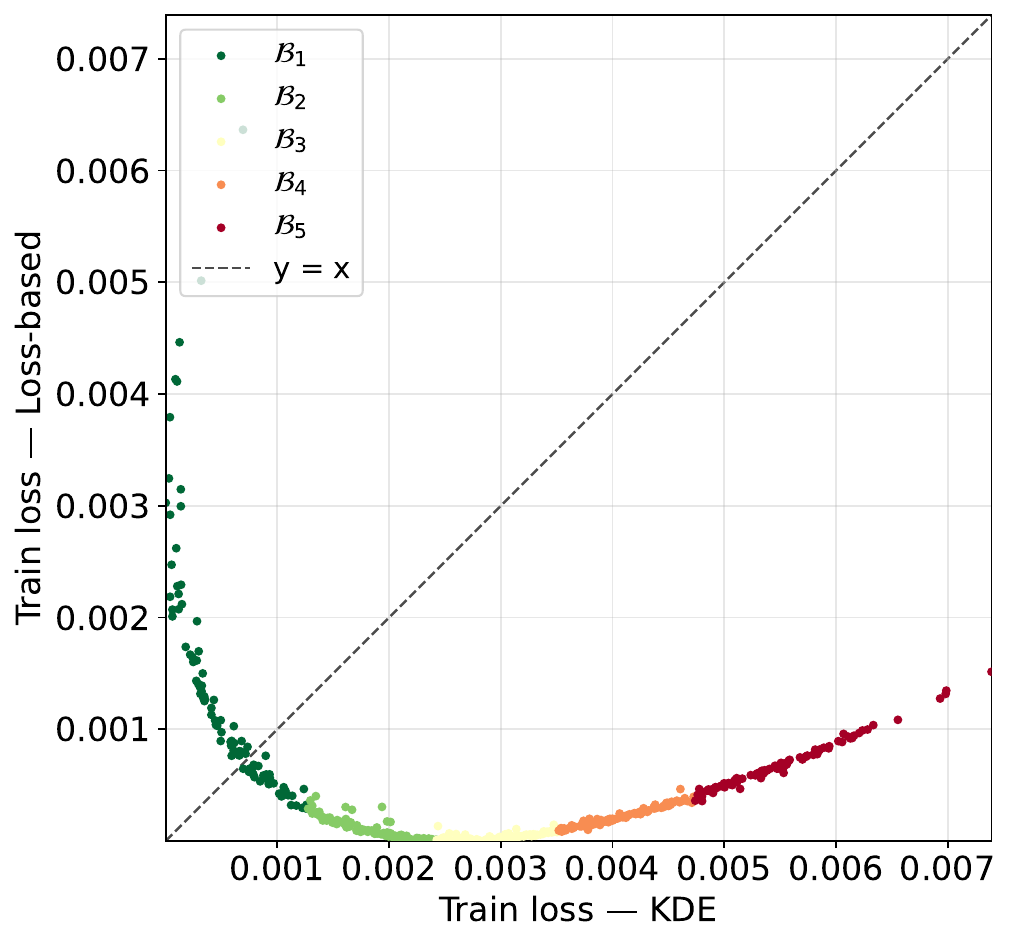}}
  \caption{\textbf{Training dynamics under different CL strategies.} (a)(b): Training curves of different strategies across buckets under the Baby-Steps schedule. Distribution of instance train losses for two dataset-strategy pairs. (c)(d)(e): Positions corresponds to instance-wise train loss in the space of strategies and coloring to the bucketing of the $x$-axis strategy. Instances located above the $y=x$ line are considered easier by the $x$-axis strategy (and conversely)}
  \label{fig:train_evol}
\end{figure}

\inlinetitle{On the density-difficulty assumption}{:}
The density-based approach relies on the assumption that high-density regions in the representation space correspond to easier instances, meaning instances whose patterns are more commonly represented in the training data and therefore statistically better supported. This heuristic is inspired by a limited but existing number of works in the literature, notably CurriculumNet~\cite{guoCurriculumNetWeaklySupervised2018}, originally formulated for noisy-label classification, and its transfer to unlabeled regression requires some care. In time-series forecasting, a dense cluster in representation space corresponds to a recurrent temporal pattern (\eg a typical weekly electricity demand profile), which is indeed easier for the model to fit due to repeated exposure. Conversely, low-density instances tend to correspond to irregular subsequences or rare regime transitions, which are typically harder for a Transformer to fit reliably at early stages of training. The empirical observation that density-based methods outperform loss-based ones on ILI and ETT provides indirect support for this heuristic in the forecasting setting.

\subsection{Insights on the training under CL strategies}

To supplement our analysis and provide more insights on the distinct effects difficulty measures may have, we provide \Fig{fig:train_evol} that explores the evolution of training of $\trainedModel$ across curriculum stages under Baby-Steps.

In (a), each learning curve is rather monotonous within each bucket and also when looking at the end state of each pair of subsequent buckets, while the training experiences -as expected- little jumps  when a new curriculum bucket is taken in.

In (b), gains from using curricula are shown for the first four buckets, but a sudden increase of the train loss occurs for the last one, both for the loss- and density-based method. This effect does not however prevent the models to arrive to overall better solutions when using CL strategies compared to random bucketing (see Table\,\ref{tab:test_losses_merged}), and may be refined with more adaptive scheduling strategies.

In the scatter plots (c-e), we compare the instance-based loss as assessed by two strategies in each case, with a coloring indicating the buckets for the strategy of the x-axis.
This visualization reveals how $\trainedModel$ perceives instance difficulty under different criteria, and how these align or not. For instance, in (c), the comparison between STL and Convex-value shows a correlated pattern, while Convex-value dominates almost everywhere, which supports its superior performance in the test set. In (d), loss- and density-based curricula compare to each other, and in this case they rather agree but there is also evidence for complementarity (\ie not perfect alignment on the diagonal). Conversely, in (e), we see a more erratic case where the loss- and the density-based (KDE) curricula are in important disagreement: easy instances for one are hard to the other, and vice versa. Interestingly, KDE yields a greater test MSE than Loss-based in this setting ($0.026$ vs $0.077$ Table~\ref{tab:test_losses_merged}), providing evidence on the value of density-based approaches.

\section{Conclusion}
In this work, we introduced DCCL, a modular framework for model-driven Curriculum Learning, leveraging both loss-based and density-based difficulty measures. Motivated by the interplay of data structure and model learning capacity, we introduced hybrid dual-criterion strategies to estimate instance difficulty. Our empirical evaluation was conducted on a Transformer over multiple time-series forecasting benchmarks, a domain that is rather unexplored in the literature. Our results demonstrate that hybrid dual-criterion strategies can effectively combine complementary insights over learning difficulty, yielding consistent improvements over single-criterion curricula and no-curriculum baselines. Future work can extend the framework to other tasks and difficulty assessment measures, and explore adaptive scheduling strategies that dynamically adjust curriculum progression based on model performance.

\begin{credits}
\subsubsection{\ackname}
A.K. and M.B. acknowledge support from the Industrial Data Analytics and Machine Learning Chair hosted at ENS Paris-Saclay, Université Paris-Saclay.%

\subsubsection{\discintname}
The authors have no competing interests to declare that are relevant to the content of this article.
\end{credits}
%
%
\bibliographystyle{splncs04}
\bibliography{CL_TS}

\newpage

\appendix

\end{document}

%% file: definitions.tex
\usepackage{xspace}
\usepackage{enumitem}

\usepackage{booktabs} 
\usepackage{multirow}

\usepackage{hyperref}
\hypersetup{
  colorlinks=true,breaklinks,
  linktoc=section,
  linkcolor=red,
  citecolor=blue,
  urlcolor=blue,
  }

\usepackage[export]{adjustbox}
\usepackage[table,dvipsnames]{xcolor}
\usepackage{tcolorbox}

\usepackage{graphicx}

\usepackage{lmodern}
\usepackage[T1]{fontenc}
\usepackage{mathtools}
\usepackage{amssymb}
\usepackage{amsfonts}
\usepackage{bbm}

\usepackage{algorithm}
\usepackage{algpseudocode}

\newcommand{\overbar}[1] {\mkern 1.5mu\overline{\mkern-3.5mu#1\mkern-1.0mu}\mkern 1.5mu}

\newcommand{\Fig}[1]		{Fig.\,\ref{#1}}

\newcommand{\Tab}[1]		{Tab.\,\ref{#1}}

\newcommand{\ie}   			{i.e.\@\xspace}
\newcommand{\eg}   			{e.g.\@\xspace}

\newcommand{\notX}[1][] {\overbar{X} \ifthenelse{\isempty{#1}}{}{_{\!#1}}} 

\newcommand{\inlinetitle}[2]  {\smallskip\noindent\textbf{\emph{#1}{#2}}}

\usepackage[normalem]{ulem}

\newcounter{marginNoteCounter}

\DeclareMathAlphabet{\altmathcal}{OMS}{cmsy}{m}{n}
\renewcommand{\mathcal}[1]  {\altmathcal{#1}}

\makeatletter
\DeclareRobustCommand*{\escapeus}[1]{%
    \begingroup\@activeus\scantokens{#1 }\endgroup}
\begingroup\lccode`\~=`\_\relax
    \lowercase{\endgroup\def\@activeus{\catcode`\_=\active \let~\_}}
\makeatother

\usepackage{scalerel}
\usepackage{tikz}
\usetikzlibrary{svg.path}
\usetikzlibrary{arrows.meta, positioning, decorations.pathreplacing, shapes.geometric, calc, fit, backgrounds}

\definecolor{orcidlogocol}{HTML}{A6CE39}
\tikzset{
  orcidlogo/.pic={
    \fill[orcidlogocol] svg{M256,128c0,70.7-57.3,128-128,128C57.3,256,0,198.7,0,128C0,57.3,57.3,0,128,0C198.7,0,256,57.3,256,128z};
    \fill[white] svg{M86.3,186.2H70.9V79.1h15.4v48.4V186.2z}
                 svg{M108.9,79.1h41.6c39.6,0,57,28.3,57,53.6c0,27.5-21.5,53.6-56.8,53.6h-41.8V79.1z M124.3,172.4h24.5c34.9,0,42.9-26.5,42.9-39.7c0-21.5-13.7-39.7-43.7-39.7h-23.7V172.4z}
                 svg{M88.7,56.8c0,5.5-4.5,10.1-10.1,10.1c-5.6,0-10.1-4.6-10.1-10.1c0-5.6,4.5-10.1,10.1-10.1C84.2,46.7,88.7,51.3,88.7,56.8z};
  }
}

\newcommand\orcidicon[1]{\href{https://orcid.org/#1}{\mbox{\scalerel*{
\begin{tikzpicture}[yscale=-1,transform shape]
\pic{orcidlogo};
\end{tikzpicture}
}{|}}}}